# Semantic Filtering by inference on domain knowledge in spoken language dialogue systems

**Afzal Ballim and Vincenzo Pallotta**

MEDIA group at the Theoretical Computer Science Laboratory (LITH)
Swiss Federal Institute of Technology - Lausanne (EPFL)

IN-F Ecublens 1015 Lausanne (Switzerland)
Phone: +41-21-693 52 97 Fax:+41-21-693 52 78
{ballim,pallotta}@di.epfl.ch

**Abstract**

General natural dialogue processing requires large amounts of domain knowledge as well as linguistic knowledge in order to ensure acceptable coverage and understanding. There are several ways of integrating lexical resources (e.g. dictionaries, thesauri) and knowledge bases or ontologies at different levels of dialogue processing. We concentrate in this paper on how to exploit domain knowledge for filtering interpretation hypotheses generated by a robust semantic parser. We use domain knowledge to semantically constrain the hypothesis space. Moreover, adding an inference mechanism allows us to complete the interpretation when information is not explicitly available. Further, we discuss briefly how this can be generalized towards a predictive natural interactive system.

## 1. Introduction

The domain we are concerned with is in interaction through speech with information systems. The availability of a large collection of annotated telephone calls for querying the Swiss phone-book database (the Swiss French PolyPhone corpus Chollet et al., 1996) allowed us to propose and evaluate a first functional prototype of a software architecture for vocal access to the database through the phone and to test our recent findings in semantic robust analysis obtained in the context of the Swiss National Fund research project ROTA (Robust Text Analysis) (Ballim and Pallotta, 2000), and in the recent Swisscom funded project ISIS (Interaction through Speech with Information Systems) (Armstrong et al., 1999). The general applicative framework of the ISIS project[1] was to design an information system NLP interface for automated telephone-based phone-book inquiry. The objective of the project was to define an architecture to improve speech recognition results by integrating higher level linguistic and domain knowledge.

One of the main issues which has been taken into consideration is about robustness. Robustness in dialogue is crucial when the artificial system takes part in the interaction since inability or low performance in processing utterances will cause unacceptable degradation of the overall system. As pointed out in (Allen et al., 1996) it is often better to have a dialogue system that tries to guess a specific interpretation in case of ambiguity rather than ask the user for a clarification. If this first commitment results later have to been a mistake, a robust behavior will be able to interpret subsequent corrections as repair procedures to be issued in order to get the intended interpretation.

### 1.1. The ISIS architecture

Dialogue processing requires in general large amounts of domain knowledge as well as linguistic knowledge in order to ensure acceptable coverage and understanding in unrestricted domains. Cooperation between processing modules and the integration of various knowledge resources require the design of a suitable software architecture. In the ISIS project the processing of the corpus data is performed at various linguistic levels performed by modules organized into a pipeline. Each module assumes as input the output produced at the previous stage.

This is not the optimal solution. We used this simple and naive architecture since one of the goals of our project is to understand how far it was possible go in natural language speech understanding without using any kind of feedback among different linguistic modules.

The proposed architecture for the functional prototype contained 3 modules:

1. a speech recognition system taking speech signals as input and providing N-best sequences in form of a lattice (Andersen 1997);

---

[1] The ISIS project started on April 1998 and finished on April 1999. It was funded and overseen by SwissCom; the partners were three Swiss institutions, namely LIA (Artificial Intelligence Laboratory) and LITH (Theoretical Computer Science Laboratory) at EPFL (Swiss Federal Institute of Technology), ISSCO (``Dalle Molle'' Institute for Semantic and Cognitive Studies) and IDIAP (``Dalle Molle'' Institute for Perceptual Artificial Intelligence).

2. a stochastic syntactic analyzer (i.e. parser) extracting the k-best analyses (Chappelier and Rajman 1998);

3. a semantic module in charge of filling the frames required to query the database (Ballim and Pallotta 1999).

It is worthwhile to remark that an improvement of the ISIS architecture has been achieved using the produced k-best syntax analyses to prune back the lattice produced by the speech recognition system. This kind of feedback has been proven to have a great impact in reducing the ambiguity induced by the speech recognizer, thus improving the overall system precision and performance. In this paper we will not concern ourselves with the above aspect, which can be found in the project's final report, but we are going to consider here only details about the functionality of the semantic module.

## 1.2. Methodology

The processing of the corpus data is performed at various linguistic levels by modules organized into a pipeline. The main goal of this architecture is to understand how far it is possible to go without using any kind of feedback and interactions between different linguistic modules. In a first stage, morphologic and syntactic processing1 are applied to the output from the speech recognizer module which usually produces a large word-graph hypothesis. Thus the forest of syntactic trees produced by this phase have been used to achieve two different goals:

1. The n-best analyses are use to disambiguate speech recognizer hypothesis

2. They served as supplementary input for the robust semantic analysis that we performed, that had as goal the production of query frames for the information system.

Although robustness can be considered as being applied at either a syntactic or semantic level, we believe it is generally at the semantic level that it is most effective. This robust analysis needs a model of the domain in which the system operates, and a way of linking this model to the lexicon used by the other components. It specifies semantic constraints that apply in the world and which allow us, for instance, to rule out incoherent requests. The degree of detail required of the domain model used by the robust analyzer depends upon the ultimate task that must be performed: in our case, furnishing a query to an information system. Taking the assumption that the information system being queried is relatively close in form to a relational database, the goal of the interpretative process is to furnish a query to the information system that can be viewed in the form of a frame with certain fields completed, the function of the querying engine being to fill in the empty fields.

The use of domain knowledge has turned out to be crucial since our particular goal is to process queries without any request of clarification from the system. Due to the inaccuracy and ambiguity generated by previous phases of analysis we need to select the best hypotheses and often recover information lost during that selection.

## 2. Semantic analysis

We proposed the use of a *light-parser* for doing sentence-level semantic interpretation and thus generate a set of ranked frame hypotheses (Ballim and Pallotta, 1999). The main idea comes from the observation that interpretation does not always need to rely on the deep structure of the sentence (e.g. at morpho-syntactic level). In some specific domains it is sometimes sufficient to find some cue-phrases which allow us to locate the logical sub-structures of the sentence. If the domain is simple enough this task can be easily mechanized.

### 2.1. Hypotheses generation

We integrate the above principle in order to effectively compute frame hypotheses for the query generation task. This can be done by building a query hypotheses lattice. The lattice of hypotheses is generated by means of a LHIP (Ballim and Russel, 1994; Ballim and Lieske, 1998) weighted grammar extracting what we called semantic chunks.

A LHIP parse may easily produce several multiple analyses. The main goal of introducing weights into LHIP rules is to induce a partial order over the generated hypotheses and exploit it for further selection of k-best analysis. The following schema illustrates how to build a simple weighted rule in a compositional fashion where the resulting weight is computed from the sub-constituents using the minimum operator. Weights are real numbers in the interval [0; 1].

```
cat(cat(Hyp),Weight) ~~>

   sub_cat1(H1,W1),
   ...,
   sub_catn(Hn,Wn),
   {app_list([H1,...,Hn],Hyp),
    min_list([W1,...,Wn],Weight)}.
```

This strategy is not the only possible since the LHIP formalism allows a greater flexibility. Without entering into formal details we can observe that if we strictly follow the above schema and we impose a strict parsing strategy then we are dealing with fuzzy DCG grammars which are in a sense similar to fuzzy context free grammars (Asveld, 1996). We actually extend this class of grammars with a notion of fuzzy-robustness where weights are used to compute confidence factors for the membership of islands to categories.

We tried to get some inspiration from the above proposal for integrating fuzzy logic and parsing to compute weights to assign to each frame filling hypotheses. Each LHIP rule returns a confidence factor together with the sequence of names. The confidence factor for a rule can be either assigned statically to pre-terminal rules (e.g. those identifying separators or introducers) or can be computed composing recursively the confidence factors of sub-constituents. Confidence factors are combined choosing the minimum among confidences of each sub-constituents. It is possible that there is no enough information for filling a slot. In this case the grammar should provide a means to provide an empty constituent when all possible hypothesis rules have failed.

At the end of this process we obtain suitable interpretations from which we are able to extract the content of the query. The rules are designed considering two kind of knowledge: domain knowledge and linguistic knowledge.

Semantic markers are domain-dependent word patterns and must be defined for a given corpus. They identify cue-phrases serving both as separators between two logical subparts of the same sentence and as anchors for semantic constituents. In our specific case they allow us to search for the content of the query only in interesting parts of the sentence. The generation of query hypotheses is performed by: composing weighted rules, assembling semantic chunks and filtering possible hypotheses.

## 2.2. Filtering

The obtained frame hypotheses can be further filtered by both using structural knowledge (e.g. constraints imposed by the syntax analysis) and domain knowledge (e.g. an ontology like Wordnet[2]). In order to combine the information extracted from the previous analysis step into the final query representation which can be directly mapped into the database query language we will make use of a frame structure in which slots represent information units or attributes in the database. A simple notion of context can be useful to fill by default those slots for which we have no explicit information. For doing this type of hierarchical reasoning we exploit the meta-programming capabilities of logic programming and we used a meta-interpreter which allows multiple inheritance among logical theories (Brogi and Turini, 1995). More precisely we made use of the special retraction operator "<" for composing logic programs which allows us to easily model the concept of inheritance in hierarchical reasoning. The expression `P` $\cup$ `Q`, where `P` and `Q` are meta-variables used to denote arbitrary logic programs, means that the resulting logic programs contains all the definition of `P` except those that are also defined in `Q`. The definition of the *isa* operator is obtained combining the retraction operator with the union operator (e.g. $\cup$) that simply making the physical union of two logic programs, by

---

[2] See WordNet 1.6 CD-rom, 1998. or
http://www.cogsci.princeton.edu/~wn

```
P isa Q = P ∪ (Q < P ):
```

As an example for the above definition we provide some default definitions which have been used to represent part of the world knowledge in our domain. The rules theory contains rules for inferring the locality or the locality type when they are not explicitly mentioned in the query.

rules:

```
locality(City) :-
caller.prefix(X), prefix(X,City).

loc.type(Type) :-
locality(City), gis(City,Type).
```

where `prefix/2` and `gis/2` are world knowledge bases (i.e. a collection of facts grouped in a theory called `kb` which wraps an existing knowledge base or ontology) and caller.`prefix/1` can be easily provided from the answer system.

If some information is missing then the system tries to provide some default additional information to complete the query. The following theory contains definition for some mandatory slots which need to be filled in case of incomplete queries, like for instance in the theory query.defaults:

query.defaults:

```
identification(person).
phone.type(standard).
loc.type(city).
```

Finally starting from an incomplete query which does not account for the required information we can use deduction to generate the query completion like for instance asking for:

```
?- demo((query isa query.default) \/
rules \/ kb), loc.type(X))
```
[3].

## 3. Summary

From a very superficial observation of the human language understanding process, it appears clear that no deep competence of the underlying structure of the spoken language is required in order to be able to process acceptably distorted utterances. On the other hand, the more experienced is the speaker, the more probable is a successful understanding of that distorted input.

In this paper we summarized a proposal for a framework for designing a knowledge-driven dialogue system. Starting with a case study and following an approach which combines the notions of robust parsing and world knowledge in sentence interpretation, we built a practical domain-dependent application. The proposed methodology can be applied whenever it is possible to superimpose a sentence-level semantic structure to a text without relying on a previous deep syntactical analysis. This kind of procedure can be also profitably used as a pre-processing tool in order to cut out part of the sentence which have been recognized to have no relevance in the understanding process in that they do not fit the system expecations.

---

[3] The predicate demo/2 is a multi-theory meta-interpreter extended for dealing with logic programs expressions. For further details refer to (Brogi and Turini, 1995).

Even if the query generation problem may not seem a critical application it should be held in mind that the sentence processing must be done on-line. Having this kind of constraints we cannot design our system without caring for efficiency and thus provide an immediate response. Another critical issue is related to whole robustness of the system. In our case study we tried to make experiences on how it is possible to deal with an unreliable and noisy input without asking the user for any repetition or clarification. This may correspond to a similar problem one may have when processing text coming from informal writing such as e-mails, news and in many cases Web pages where it is often the case to have irrelevant surrounding information.

## 4. Discussion

So far we have presented a robust speech understanding system that is not far removed from many other systems. In particular, keyword spotting is a technique often used in restricted domains. Certainly, we go further by using weighting techniques on the grammar, employing a logical intermediate representation, performing inference on this intermediate representation, and thus filling the template. The question we now wish to address, is how can we move forward. Can this approach be generalized? What are the consequences of this approach? We will argue that this method fits into a general approach that we call a predictive dialogue modelling approach. First, however, it is necessary to mixed in general remarks about the state of the art in dialogue processing and the problems that must be addressed.

The advancement from system directed queries to mixed strategies is an important first stage in allowing for more natural interactive systems. Of course, a mixed initiative approach typically generates higher error rates. Reducing these error rates involves constraining dialogues which is typically done by restricting the domain of application of the system. Such an approach allows us to restrict the vocabulary to maybe a few hundred words instead of the thousands or hundreds of thousands of words that we would need in a more general case. An observation of human to human communication shows a large number of phenomena which present particular problems for machine analysis. Interruptions, confirmations, anaphora, ellipsis as well as the breaks, repairs, pauses, and jumps normally found in human dialogue all present difficulties for machine understanding. Robust processing goes a long way to handling certain of these problems. We contend, however, that more general solutions can only come from having a model of the domain and of the user.

The model of the user is not only necessary for better understanding what the user is saying, but also for matching the expectations of the user in the interaction with the machine. This is necessary because it is difficult to communicate the system's capabilities to the user. The user does not necessarily know the vocabulary that the system's capable of handling, nor the type of questions that the system may answer.

We can see then that a user model can be of great benefit in future natural interactive systems. In addition, in multimodal interaction the user model will allow us to better tailor the use of different modalities to the user. More importantly, from our point of view, such a model is part of a predictive approach to natural interactivity.

The idea of this approach is to continuously anticipate the interaction with the user. In other words, analysis should be based on the expectations of the system. Such an approach allows us to restrict vocabulary, domain knowledge, and interaction types to only those necessary for the immediate understanding. In a sense dialogue grammars, finite state approaches to dialogue, and template approaches to dialogue are all predictive models. We anticipate an approach in which more general models of language, based on the content of communication, are derived from knowledge of the domain, the user's knowledge of the domain, and the system's view of the user's needs, beliefs, goals and motivations.

### 4.1. Related works

As examples of robust approaches applied to dialogue systems we cite here two systems which are based on similar principles.

In the DIALOGOS human-machine telephone system (see Albesano et al., 1997) the robust behavior of the dialogue management module is based both on a contextual knowledge base of pragmatic-based expectations and the dialogue history. The system identifies discrepancies between expectations and the actual user behavior and in that case it tries to rebuild the dialogue consistency. Since both the domain of discourse and the user's goals (e.g. railway timetable inquiry) are clear, it is assumed the systems and the users cooperate in achieving reciprocal understanding. Under this underlying assumption the system pro-actively asks for the query parameters and it is able to account for those spontaneously proposed by the user.

In the SYSLID project (Boros et al., 1996) where a robust parser constitutes the linguistic component of the query-answering dialogue system. An utterance is analyzed while at the same time its semantical representation is constructed. This semantical representation is further analyzed by the dialogue control module which then builds the database query. Starting from a word graph generated by the speech recognizer module, the robust parser will produce a search path into the word graph. If no complete path can be found, the robust component of the parser, which is an island based chart parser (Hanrieder and Goerz, 1995), will select the maximal consistent partial results. In this case the parsing process is also guided by a lexical semantic knowledge base component that helps the parse in solving structural ambiguities.

### 4.2. Future Work

The limited resources of the project did not allow us to adequately evaluate the results and test the system against real situations. Nonetheless our final opinion about the ISIS project is that there are some promising directions applying robust parsing techniques and integrating them with knowledge representation and reasoning. Moreover we did not commit on the used architecture and we envision that better results can be achieved moving towards a distributed agent-based architecture for natural language processing. An ongoing project[4] at our laboratory is concerned with these aspects, where we propose an hybrid distributed architecture which combines symbolic and numerical computing by means of agents providing linguistic services. Within this architecture also the knowledge management plays a central role and it is aimed to the intelligent coordination of the linguistic agents (Ballim and Wilks, 1991; Ballim, 1993).

Another importat aspect we are certainly interested in taking into account is related to multi-modal interaction. Considering for instance, prosodic information a more robust and efficient dialogue system can be obtained as shown in (Kompe et al., 1994) in the context of the VERBMOBIL project (Jekat et al., 1995).

## 5. References

Albesano, Dario, Paolo Baggia, Morena Danieli, Roberto Gemello, Elisabetta Gerbino and Claudio Rullent. DIALOGOS: a robust system for human-machine spoken dialogue on the telephone. In Proc. of ICASSP, Munich, Germany, 1997.

Allen, J.F., B. Miller, E. Ringger, and T. Sikorski. A robust system for natural spoken dialogue. In Proc. 34th Meeting of the Assoc. for Computational Linguistics. Association of Computational Linguistics, June 1996.

Andersen, J.M., G. Caloz, and H. Bourlard. Swisscom "advanced vocal interfaces services" project. Technical Report COM-97-06, IDIAP, Martigny, December 1997.

Armstrong, S., A. Ballim, P. Bouillon, J.-C. Chappelier, V. Pallotta and M. Rajman, 1999 ISIS project: final report. Technical report, Computer Science Department – Swiss Federal Institute of Technology, September 1999. ftp://liaftp.epfl.ch/lia/chaps/Isis/rapportfinal.ps.gz.

Asveld, P.R.J., 1996. Towards robustness in parsing - fuzzifying context-free language recognition. In J. Dassow, G. Rozemberg, and A. Salomaa, editors, Developments in Language Theory II - At the Crossroad of Mathematics, Computer Science and Biology, pages 443-453. World Scientific, Singapore, 1996.

Ballim, A. & Y. Wilks (1991) "Artificial Believers", Lawrence Erlbaum Associates, Hillsdale, New Jersey.

Ballim, A. (1993) "Propositional Attitude Framework Requirements," Journal for Experimental and Theoretical Artificial Intelligence (JETAI) 5, 89-100.

Ballim, A. & Y. Wilks (1991) "Beliefs, Stereotypes and Dynamic Agent Modelling," User Modelling and User-Adapted Interaction 1 (1), 33-65.

Ballim, A. and G. Russell. LHIP: Extended DCGs for Configurable Robust Parsing. In Proceedings of the 15th International Conference on Computational Linguistics, pages 501 - 507, Kyoto, Japan, 1994. ACL.

Ballim, A. and C. Lieske, 1998. Rethinking natural language processing with PROLOG. In Proceedings of the Practical Applications of PROLOG and Practical Applications of Constraint Technology (PAPPACT98), London, UK, 1998. Practical Application Company.

Ballim, A. and V. Pallotta, 1999. Robust Parsing techniques for semantic analysis of natural language queries. In Rodolfo del Monte, editor, Proceedings of (VEXTAL99) international conference, November 1999, Venezia, Italia.

Ballim, A. and V. Pallotta, 2000. Extended definite clause grammars for robust text analysis. Technical report, Swiss National Science Foundation, February 2000. http://lithwww.epAE.ch/~pallotta/rota.ps.gz.

Boros, Manuela, Gerhard Hanrieder, and Ulla Ackermann. Linguistic processing for spoken dialogue systems - experiences made in the SYSLID project -. In Proceedings of the third CRIM-FORWISS Workshop, Montreal, Canada, 1996.

Brogi, A. and F. Turini, 1995 Meta-logic for program composition: Semantic issues. In K.R. Apt and F.Turini, editors, Meta-Logics and Logic Programming. The MIT Press.

Chappelier, J-C. and M. Rajman, 1998. A generalized CYK algorithm for parsing stochastic Context Free Grammars. In 1st Workshop on Tabulation in Parsing and Deduction (TAPD98), pages 133-137, Paris, April 2-3 1998.

---

[4] More information about the HERALD (Hybrid Environment for Robust Analysis of Language Data) project are available at: http://lithwww.epfl.ch/~pallotta/herald.ps.gz

Chollet, C., J.-L. Chochard, A. Constantinescu, C. Jaboulet and Ph. Langlais, 1996. Swiss french polyphone and polyvar: Telephone speech database to model inter- and intraspeaker variability. Technical Report RR-96-01, IDIAP, Martigny, April 1996.

Hanrieder, G., and G. Goerz. Robust parsing of spoken dialogue using contextual knowledge and recognition probabilities. In Proceedings of the ESCA Tutorial and Research Workshop on Spoken Dialogue Systems * Theories and Applications, pages 57-60, Denmark, May 1995.

Jekat, S., A. Klein, E. Maier, I. Maleck, M. Mast, and J.J. Quantz. Dialogue acts in vermobil. Verbmobil Report 65, DFKI, 1995.

Kompe, R., A. Kiebling, T. Kuhn, M. Mast, H. Niemann, E. Noeth, K. Ott, and A. Batliner. Prosody takes over: A prosodically guided dialog system. Verbmobil report 47, DFKI, 1994.